\documentclass[letterpaper,10pt]{article} 

\usepackage{opticameet3} 
\usepackage{wrapfig}
\usepackage{float}
\usepackage{bm}
\usepackage{amsmath}
\usepackage{amsfonts}
\usepackage{amssymb}
\usepackage{calligra}
\usepackage{calrsfs}
\usepackage{mathrsfs}
\usepackage{mathtools}
\usepackage{amsthm}
\usepackage[margin=1cm]{caption}
\newcommand\authormark[1]{\textsuperscript{#1}}

\usepackage{amsmath,amssymb}
\usepackage[colorlinks=true,bookmarks=false,citecolor=blue,urlcolor=blue]{hyperref} 

\begin{document}

\title{Experimental Demonstration of an Optical Neural PDE Solver via On-Chip PINN Training}


\author{Yequan Zhao,\authormark{1,2} Xian Xiao,\authormark{1} Antoine Descos,\authormark{1} Yuan Yuan,\authormark{1} Xinling Yu,\authormark{1,2} Geza Kurczveil,\authormark{1} Marco Fiorentino,\authormark{1} Zheng Zhang\authormark{2} and Raymond G. Beausoleil\authormark{1}}

\address{\authormark{1} Hewlett Packard Labs, Hewlett Packard Enterprise, 820 N. McCarthy Blvd., Milpitas, California 95305, USA\\
\authormark{2} Department of Electrical and Computer Engineering, University of California, Santa Barbara, CA 93106, USA}

\email{yequan\_zhao@ucsb.edu, xian.xiao@hpe.com} 

\begin{abstract}
Partial differential equation (PDE) is an important math tool in science and engineering. This paper experimentally demonstrates an optical neural PDE solver by leveraging the back-propagation-free on-photonic-chip training of physics-informed neural networks.
\end{abstract}

\begin{figure}[!b]
    \vspace{-15pt}
    \centering
    \includegraphics[width=\linewidth]{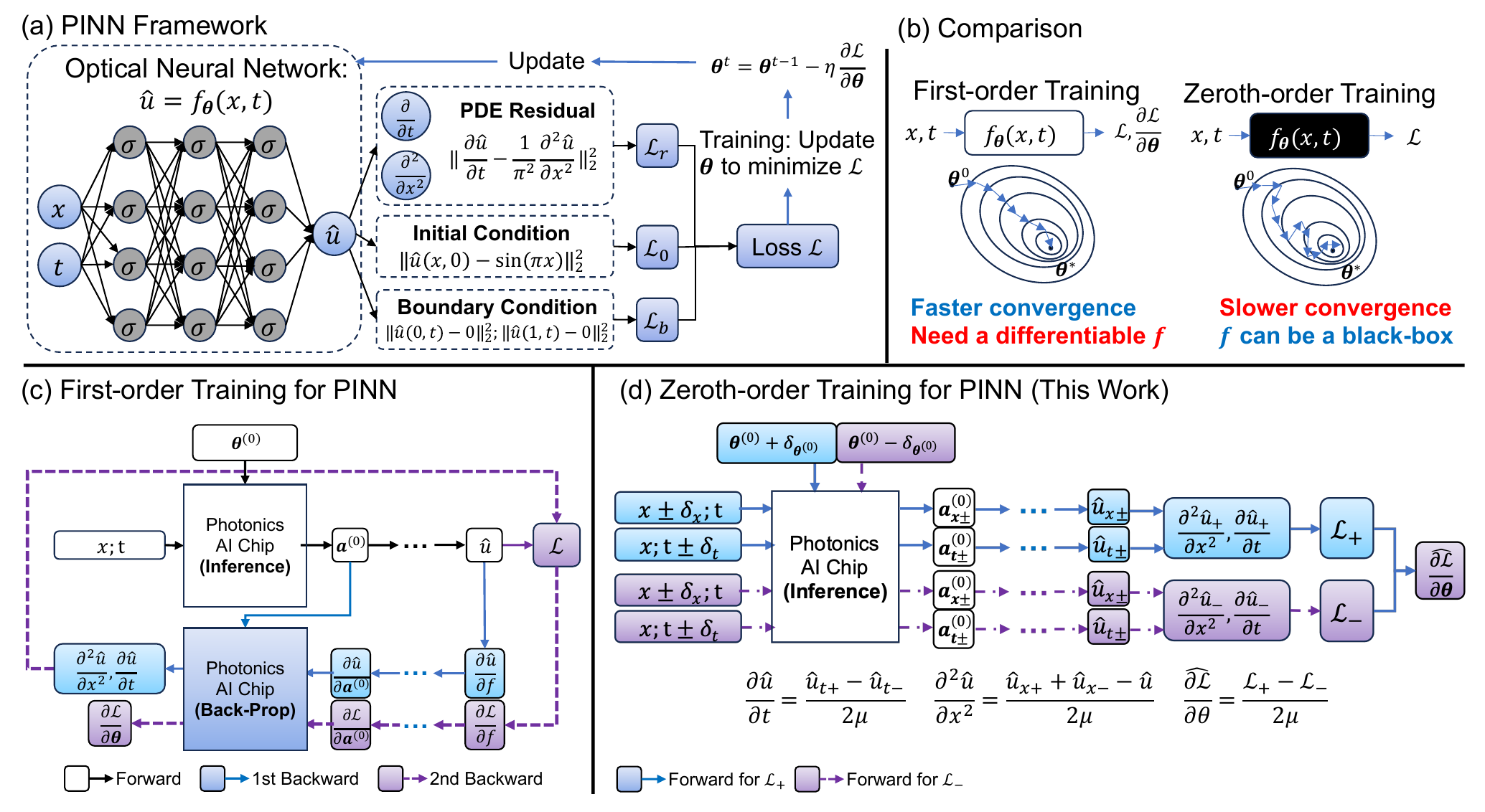}
    \vspace{-25pt}
    \captionsetup{width=\textwidth, font={small},justification=raggedright}
    \caption{(a) PINN framework. (b) Comparison between first-order training and zeroth-order training. (c) First-order training framework of PINN.
    (d) Zeroth-order training framework of PINN proposed in this work.}
    \label{fig:Overall_architecture}
    \vspace{-20pt}
\end{figure}

\section{Introduction}

Partial differential equation (PDE) is one of the most important math tools in science and engineering. Examples include electromagnetic modeling and thermal analysis of IC chips~\cite{li2004efficient}, medical imaging~\cite{villena2015marie}, safety verification of autonomous systems~\cite{bansal2021deepreach}. Discretization-based solvers (e.g., finite-difference and finite-element methods) convert a PDE to a large-scale algebraic equation via spatial and temporal discretization. Solving the resulting algebraic equation often requires massive digital resources and run times. Physics-informed neural network (PINN) is a promising discretization-free and unsupervised approach to solve PDEs~\cite{Raissi}. PINN uses the residuals of a PDE operator and the boundary/initial conditions to set up a loss function, then minimizes the loss to train a neural network as a global approximation of the PDE solution. 
However, current PINN training typically needs several to dozens of hours on a powerful GPU, hindering the deployment of an real-time neural PDE solver on edge devices.


Optical neural networks (ONNs) provide a promising high-throughput, low-energy-consumption, low-latency, and high-parallelism solution. However, training PINNs on a photonic chip is very challenging. It is hard to realize training on photonic chips with back propagation (BP) due to 1) BP needs extra memory and photonics hardware to implement backward computation graphs and 2) BP needs a fully differentiable neural network, requiring a calibration process to accurately model the on-chip noises and imperfections. Several BP-free~\cite{ref:GuDAC} and in-situ BP methods~\cite{ref:Pai} are proposed, but they all have poor scalability. This problem becomes more severe in PINN training since the loss function also includes derivative terms, requiring multiple BPs.

In this paper, we experimentally demonstrate an optical neural PDE solver by calibration-free and BP-free on-chip training of PINNs. 
We leverage a photonics-friendly back-propagation-free training algorithm~\cite{zhao2023real} and implement it on a photonics AI chip (a 1×4 micro-ring resonator (MRR) weight bank). The on-chip training experiment learns a solution for a one-dimensional heat equation with 5E-3 $\ell_2$ error after 1000 iterations of update, which experimentally validates our concepts of training PINN on an optical chip to solve a PDE.


\vspace{-5pt}
\section{Principle and Architecture}
\vspace{-5pt}

Fig.~\ref{fig:Overall_architecture}(a) shows the PINN architecture. 
A PINN is a feed-forward neural network parameterized by $\bm{\theta}$ that uses the output $\hat{{u}}({x}, t)=f_{\bm{\theta}}({x}, t)$ to approximate the PDE solution ${u}({x}, t)$, where ${x}$ and $t$ repersent the spatial and temporal dimensions, respectively. 
We give an example of training a PINN to solve a heat equation with one (1) spatial dimension $(x)$ and one (1) temporal dimension $(t)$: $u_t=\frac{1}{\pi^2} u_{xx}$. The initial condition is $u(x,0)=\sin (\pi x)$, and boundary conditions are $u(0,t)=0;~ u(1,t)=0$.
During training, $\bm{\theta}$ are updated to minimize a physics-informed loss function $\mathcal{L}=\mathcal{L}_r+\mathcal{L}_0+\mathcal{L}_b$. 
$\mathcal{L}_r, \mathcal{L}_0, \mathcal{L}_b$ measure how well the approximation $\hat{u}$ complies with the PDE operator, the initial conditions, and boundary conditions that describe the physics laws of a system, respectively. 
Fig.~\ref{fig:Overall_architecture}(b) gives a comparison between first-order training and zeroth-order training. First-order training has a faster convergence rate since it utilizes the gradient information to guide the parameter update direction. However, as shown in Fig.~\ref{fig:Overall_architecture}(c), implementing first-order training on photonics hardware to train a PINN requires 1) additional photonics chip design for back propagation computation, which poses expensive hardware overhead and 2) an accurate and differentiable model of photonics devices, which requires an exhaustive calibration process to model the fabrication error. Moreover, since unknown environmental noises always exist, such modeling is not ideally accurate.
To address the challenge of implementing PINN training on photonics hardware, we design our BP-free training with zeroth-order (ZO) optimization (Fig.~\ref{fig:Overall_architecture}(d)). 
The training uses forward propagation only by repeatedly calling a photonics inference accelerator.
Since zeroth-order training does not require a differentiable neural network model, we can directly optimize the tunable parameters of photonic devices (\textit{e.g.,} voltage values of a micro ring resonators, phase shifter values of Mach-Zehnder interferometers) to minimize the training loss. 
The fabrication errors and other unknown environmental noises can be mitigated on-the-run, thus a pre-calibration process is not required.

\begin{figure}[t]
\centering
\includegraphics[width=\textwidth]{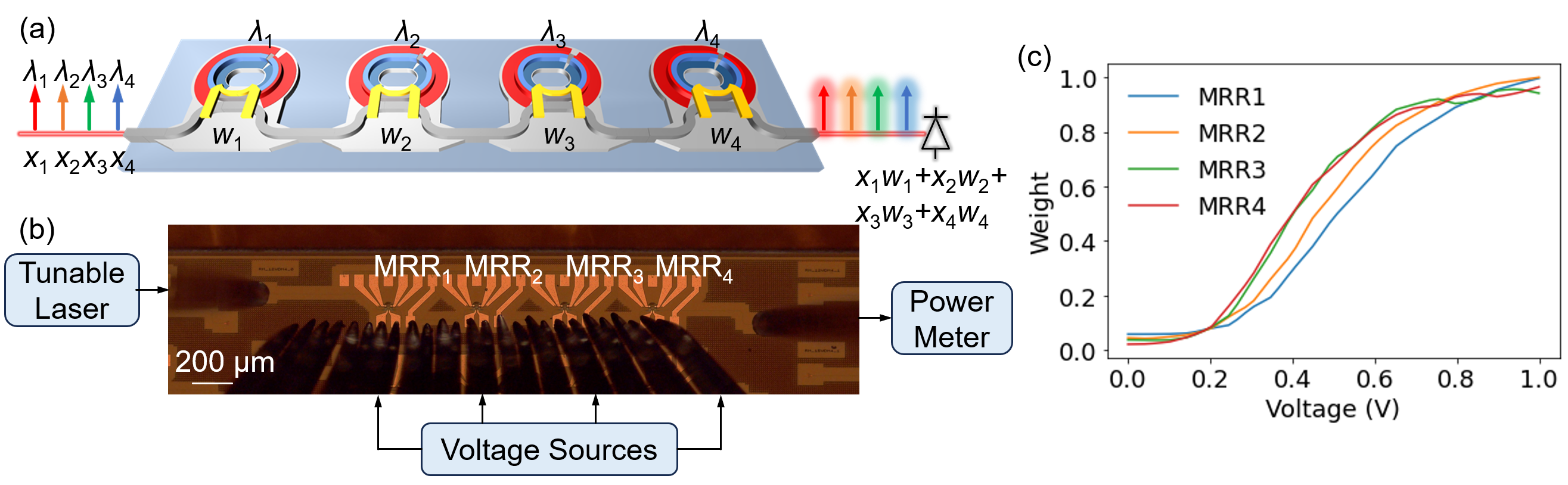}
\vspace{-25pt}
\captionsetup{width=\textwidth, font={small},justification=raggedright}
\caption{(a) Schematic of the 1$\times$4 MRR weight bank. (b) Microscope image of the weight bank and the schematic of the experimental setup for on-chip training. (c) Weight curves of the four MRRs.}
\label{fig:setup}
\vspace{-20pt}
\end{figure}

\vspace{-5pt}
\section{Experimental Demonstration}
\vspace{-5pt}

We experimentally demonstrated the optical neural PDE solver for the heat equation by training a small-scale PINN. The PINN contains a feed-forward neural network with two hidden layers, as shown in Fig.~\ref{fig:Overall_architecture}(a). The scale of the weight matrices are 2$\times$4, 4$\times$4, 4$\times$4, and 4$\times$1, respectively. The weight matrices are computed by 1$\times$4 tiles in different clock cycles with time multiplexing. The 1$\times$4 tile is implemented by a 1$\times$4 microring resonator (MRR) weight bank, as shown in Fig.~\ref{fig:setup}(a). Using the wavelength division multiplexing (WDM) technology, by encoding the input data at $\lambda_1, \lambda_2, \lambda_3, \lambda_4$, the data can multiply with each of the MRR weights $w_1, w_2, w_3, w_4$, and a photodetector at output can compute $\lambda_1w_1+\lambda_2w_2+\lambda_3w_3+\lambda_4w_4$. Fig.~\ref{fig:setup}(b) shows the experimental setup and the microscope image of the MRR weight bank chip fabricated at AMF foundry. Fig.~\ref{fig:setup}(c) shows the weight curves with different thermal tuning voltages, indicating the tuning range of the weight values from 0 to 1. The variances of the weight curves among different MRRs caused by fabrication errors can be mitigated by our BP-free on-chip training without the need for pre-calibration.

\begin{figure}[t]
\centering
\includegraphics[width=\textwidth]{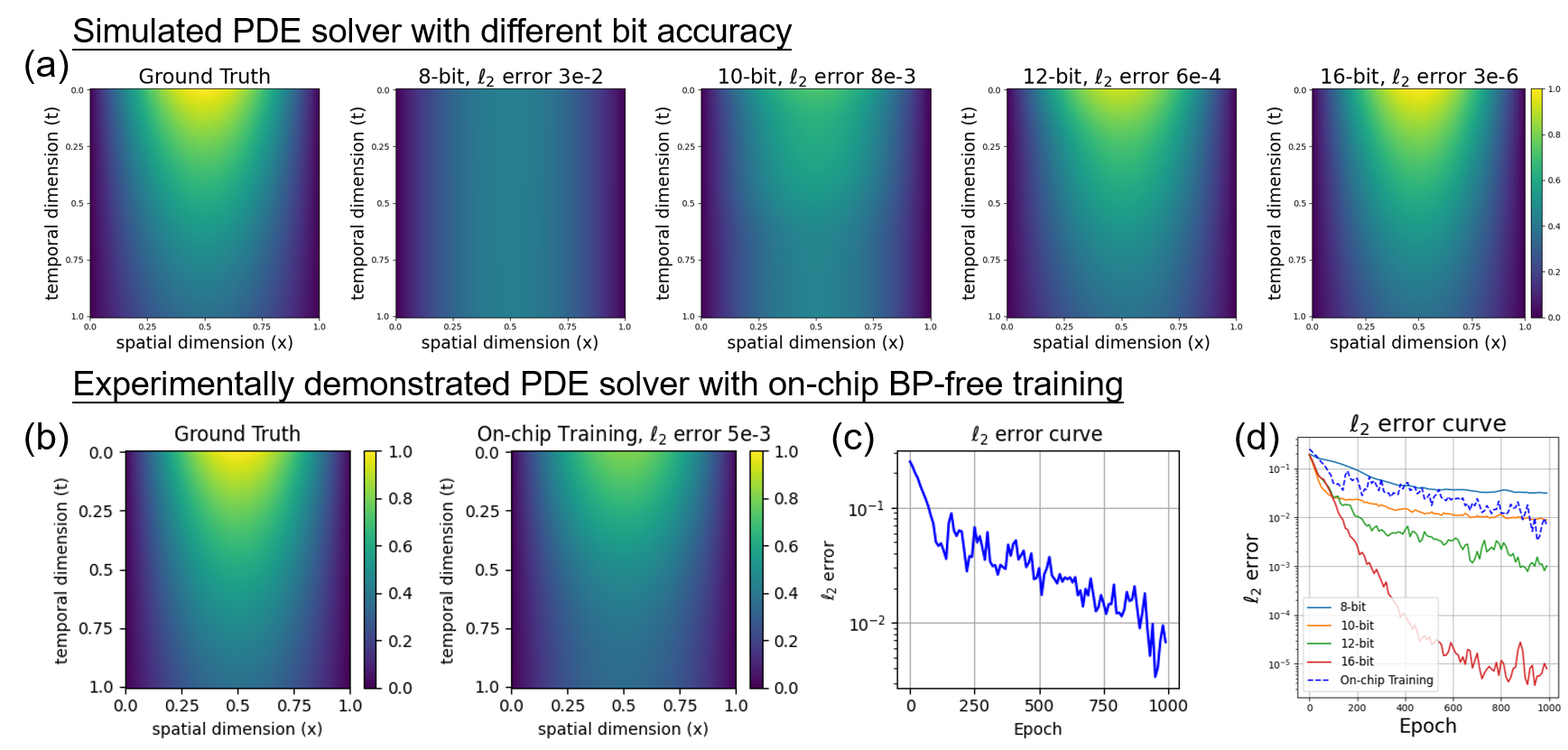}
\vspace{-25pt}
\captionsetup{width=\textwidth, font={small},justification=raggedright}
\caption{Comparison between the ground truth solution and learned solutions of (a) BP-free training simulation and (b) experimentally demonstrated BP-free on-chip training. (c) The $\ell_2$ error curve of on-chip training. (d) The $\ell_2$ error curve comparison between on-chip training and different simulation setups.}
\label{fig:Results}
\vspace{-25pt}
\end{figure}

{\bf Simulation Results.} We first numerically simulate the BP-free on-chip training under different bit precisions. We use an experimentally measured voltage-weight look-up table [as shown in Fig.~\ref{fig:setup}(c)] to model the MRR weight banks. The learned PDE solution is compared with the ground-truth solution by the $\ell_2$ error on a hold-out test point set unseen during training. The solutions are visualized with heatmaps. Each pixel indicates the value of the PDE solution ${u}(x, t)$. Fig.~\ref{fig:Results}(a) visualizes the ground truth solution and learned solutions. With a low bit accuracy (8-bit), the neural network cannot learn the governing physics due to the restricted expressive power. The results indicate the affect of bit accuracy on the performance of on-chip PINN training.

{\bf Hardware Demo.} Then, we experimentally demonstrated our BP-free training on the fabricated photonic chip. A tunable laser and photodetector are used to measure the spectrum of the MRR weight bank at different tuning voltages, as shown in Fig.~\ref{fig:setup}(b). Four voltage sources are used for thermally tuning the MRRs. The weight values are read out at the four resonances of the MRRs. Fig.~\ref{fig:Results}(b) visualizes the learned solution after on-chip training, and Fig.~\ref{fig:Results}(c) shows the $\ell_2$ error curve. The decreasing $\ell_2$ error shows that on-chip training gradually captures the governing physics laws described by PDE even under fabrication errors, environmental noise, etc. However, the limited bit accuracy of analog computation limits the accuracy of the final learned solution. Fig.~\ref{fig:Results}(d) shows the $\ell_2$ error curve comparison between on-chip training and simulation results with different bit accuracy setups. The experimentally demonstrated on-chip training showcases the bit accuracy between 8-bits and 10-bits.

\vspace{-10pt}
\section{Conclusion}
\vspace{-5pt}

In this paper, we have experimentally demonstrated an optical neural PDE solver for solving heat equations, using BP-free on-chip training of PINNs. The experiment results learn the solution with 5E-3 $\ell_2$ error. To scale up the optical on-chip training framework for real-size PINN (\textit{e.g.,} 1000 neurons per layer), the tensor-train decomposed PINN (TT-PINN)~\cite{TTPINN, TTPINN-ZO} implemented on an energy-efficient tensorized ONN (TONN)~\cite{Xiao} inference accelerator is a promising approach. 

\end{document}